\documentclass[10pt, a4paper]{article}
\usepackage[table]{xcolor}

\usepackage{booktabs,array}
\usepackage{inconsolata}
\definecolor{sim1}{HTML}{EAF2FB} % 0.80–0.83
\definecolor{sim2}{HTML}{D6E6F7} % 0.83–0.86
\definecolor{sim3}{HTML}{B8D4F2} % 0.86–0.89
\definecolor{sim4}{HTML}{90BDEB} % 0.89–0.91
\definecolor{sim5}{HTML}{5EA0E6} % 0.91–0.93

\newcommand{\scorecell}[1]{%
  \begingroup\def\val{#1}%
  \ifdim \val pt<0.83pt \cellcolor{sim1}\else
  \ifdim \val pt<0.86pt \cellcolor{sim2}\else
  \ifdim \val pt<0.89pt \cellcolor{sim3}\else
  \ifdim \val pt<0.91pt \cellcolor{sim4}\else
                         \cellcolor{sim5}\fi\fi\fi\fi
  \val\endgroup
}

\definecolor{c1}{HTML}{FFFFD9} % <30
\definecolor{c2}{HTML}{EDF8B1} % 30–35
\definecolor{c3}{HTML}{C7E9B4} % 35–40
\definecolor{c4}{HTML}{7FCDBB} % 40–45
\definecolor{c5}{HTML}{41B6C4} % 45–50
\definecolor{c6}{HTML}{1D91C0} % 50–55
\definecolor{c7}{HTML}{225EA8} % 55–60
\definecolor{c8}{HTML}{0C2C84} % ≥60
\definecolor{diaggray}{RGB}{235,235,235}
\definecolor{headcol}{RGB}{225,235,245}

\newcommand{\acc}[1]{%
  \begingroup\def\val{#1}%
  \ifdim \val pt<30pt \cellcolor{c1}\else
  \ifdim \val pt<35pt \cellcolor{c2}\else
  \ifdim \val pt<40pt \cellcolor{c3}\else
  \ifdim \val pt<45pt \cellcolor{c4}\else
  \ifdim \val pt<50pt \cellcolor{c5}\else
  \ifdim \val pt<55pt \cellcolor{c6}\else
  \ifdim \val pt<60pt \cellcolor{c7}\else
                    \cellcolor{c8}\fi\fi\fi\fi\fi\fi\fi
  \val\endgroup
}
\newcommand{\diag}{\cellcolor{diaggray}{--}}
\newcommand{\cardtitle}[1]{\vspace{4pt}\textbf{#1}\par\vspace{4pt}}

\usepackage{tikz}
\usepackage{subcaption}
\usetikzlibrary{calc}

\definecolor{diaggray}{RGB}{230,230,230}
\definecolor{gridline}{RGB}{210,210,210}
\colorlet{b1}{blue!10}
\colorlet{b2}{blue!20}
\colorlet{b3}{blue!30}
\colorlet{b4}{blue!40}
\colorlet{b5}{blue!50}
\colorlet{b6}{blue!60}
\colorlet{b7}{blue!70}
\colorlet{b8}{blue!80}

\newcommand{\scorecolor}[1]{%
  \ifdim #1pt<30pt \def\cellfill{b1}\else
  \ifdim #1pt<35pt \def\cellfill{b2}\else
  \ifdim #1pt<40pt \def\cellfill{b3}\else
  \ifdim #1pt<45pt \def\cellfill{b4}\else
  \ifdim #1pt<50pt \def\cellfill{b5}\else
  \ifdim #1pt<55pt \def\cellfill{b6}\else
  \ifdim #1pt<60pt \def\cellfill{b7}\else
                     \def\cellfill{b8}\fi\fi\fi\fi\fi\fi\fi
}

\usepackage[final]{lrec2026} % this is the new style
\usepackage{tcolorbox,xcolor}
\tcbuselibrary{skins,breakable}

\definecolor{entailbg}{RGB}{235,248,235}       % soft green
\definecolor{contrabg}{RGB}{253,235,235}       % soft red
\definecolor{neutralbg}{RGB}{255,249,230}      % soft yellow
\definecolor{cardborder}{RGB}{200,200,200}     % light gray border

\newtcolorbox{nlirelation}[2][]{enhanced,
  colback=#2!80!white,
  colframe=cardborder,
  boxrule=0.4pt,
  arc=3pt,
  outer arc=3pt,
  boxsep=5pt,
  left=6pt,
  right=6pt,
  top=6pt,
  bottom=6pt,
  fonttitle=\bfseries,
  title=#1,
  breakable}

\usepackage{tcolorbox,xcolor}
\tcbuselibrary{skins}

\definecolor{entailbg}{RGB}{235,248,235}       % soft green
\definecolor{contrabg}{RGB}{253,235,235}       % soft red
\definecolor{neutralbg}{RGB}{255,249,230}      % soft yellow
\definecolor{cardborder}{RGB}{200,200,200}     % light border

\newtcolorbox{relcard}[2][]{enhanced,
  colback=#2!90!white,
  colframe=cardborder,
  boxrule=0.5pt,
  arc=3pt,
  outer arc=3pt,
  boxsep=4pt,
  left=6pt,
  right=6pt,
  top=6pt,
  bottom=6pt,
  fonttitle=\bfseries,
  title=#1}

\usepackage{times}
\usepackage{latexsym}
\usepackage{caption}
\usepackage{subcaption}
\usepackage{booktabs}
\usepackage{geometry}
\usepackage{booktabs}
\usepackage{multirow}
\usepackage{array}

\usepackage{mdframed,xcolor}
\definecolor{exbg}{RGB}{247,249,252}
\definecolor{exborder}{RGB}{200,210,225}

\newmdenv[
  skipabove=6pt,
  skipbelow=6pt,
  linewidth=0.6pt,
  linecolor=exborder,
  backgroundcolor=exbg,
  roundcorner=4pt,
  innerleftmargin=8pt,
  innerrightmargin=8pt
]{relationbox}

\usepackage[T1]{fontenc}

\usepackage[utf8]{inputenc}

\usepackage{tcolorbox,xcolor}
\tcbuselibrary{skins,breakable}

\definecolor{promptbg}{RGB}{247,249,252}      % very light gray-blue
\definecolor{promptborder}{RGB}{200,210,225}  % soft border
\definecolor{prompttitle}{RGB}{0,102,204}     % deep ACL blue

\newtcolorbox{promptbox}[1][]{enhanced,
  colback=promptbg,
  colframe=promptborder,
  coltitle=white,
  colbacktitle=prompttitle,
  fonttitle=\bfseries,
  title=#1,
  boxrule=0.5pt,
  arc=3pt,
  outer arc=3pt,
  boxsep=5pt,
  top=6pt,
  bottom=6pt,
  left=8pt,
  right=8pt,
  breakable}

\usepackage{mdframed,xcolor} % in preamble

\definecolor{exbg}{RGB}{247,249,252}     % very light cool gray-blue
\definecolor{exborder}{RGB}{200,210,225} % muted blue-gray border

\newmdenv[
  skipabove=6pt,
  skipbelow=6pt,
  linewidth=0.6pt,
  linecolor=exborder,
  backgroundcolor=exbg,
  roundcorner=4pt,
  innerleftmargin=10pt,
  innerrightmargin=10pt,
  innertopmargin=6pt,
  innerbottommargin=6pt
]{examplebox}

\usepackage{microtype}

\usepackage{inconsolata}

\usepackage{graphicx}

\title{Evaluating Multilingual and Code-Switched Alignment in LLMs via
Synthetic Natural Language Inference}

\name{Samir Abdaljalil\textsuperscript{*}, Erchin Serpedin\textsuperscript{*}, Khalid Qaraqe\textsuperscript{†}, Hasan Kurban\textsuperscript{†}}

\address{\textsuperscript{*}Texas A\&M University, College Station, TX., USA \\
\textsuperscript{†}Hamad Bin Khalifa University, Doha, Qatar\\
         sabdaljalil@tamu.edu, hkurban@hbku.edu.qa\\}
\abstract{
Large language models (LLMs) are increasingly applied in multilingual contexts, yet their capacity for consistent, logically grounded alignment across languages remains underexplored. We present a controlled evaluation framework for multilingual natural language inference (NLI) that generates synthetic, logic-based premise–hypothesis pairs and translates them into a typologically diverse set of languages. This design enables precise control over semantic relations and allows testing in both monolingual and mixed-language (code-switched) conditions. Surprisingly, code-switching does not degrade, and can even improve, performance, suggesting that translation-induced lexical variation may serve as a regularization signal. We validate semantic preservation through embedding-based similarity analyses and cross-lingual alignment visualizations, confirming the fidelity of translated pairs. Our findings expose both the potential and the brittleness of current LLM cross-lingual reasoning, and identify code-switching as a promising lever for improving multilingual robustness. Code can be accessed at: \url{https://github.com/KurbanIntelligenceLab/nli-stress-testing}
 \\ \newline \Keywords{Large Language Models (LLMs), Natural Language Inference (NLI), Multilingual Alignment} }

\begin{document}

\maketitleabstract

\section{Introduction}
\label{sec:intro}
NLI \cite{dagan2005}—determining whether a \textit{hypothesis} is entailed by, contradicts, or is neutral with respect to a \textit{premise}—is a core benchmark for natural language understanding \cite{havaldar2025entailedlinesincorporatingimplication, yudanto-etal-2024-climate, mor-lan-levi-2024-exploring}. Its emphasis on fine-grained semantic distinctions has long made it a proxy for testing models’ capacity for deep reasoning \cite{cosma-etal-2024-hard}.
With LLMs, NLI has become a key tool for assessing generalization, reasoning, and knowledge encoding \cite{cheng-etal-2025-neutralizing}. Yet evaluations remain concentrated on high-resource languages—especially English—and are often embedded within downstream tasks such as QA or summarization, limiting insight into whether inference capabilities transfer consistently across languages under controlled semantic conditions.

We address this gap with a synthetic multilingual NLI framework that stress-tests cross-lingual semantic alignment via deterministic, logic-based templates encoding entailment, contradiction, and neutrality. The approach decouples logical structure from lexical and cultural priors, avoiding annotation noise and enabling direct, large-scale evaluation. Our contributions are: (1) a logic-driven method for generating synthetic multilingual NLI datasets with precise control over inference types and linguistic variation; (2) an automated evaluation protocol for measuring cross-lingual consistency in LLM semantic judgments; and (3) empirical evidence, across multiple models and languages, of systematic weaknesses in multilingual alignment.

By disentangling logical reasoning from linguistic noise, our framework offers a principled, reproducible basis for evaluating semantic alignment in multilingual LLMs. Section~\ref{sec:related} reviews related work, Section~\ref{sec:method} details the methodology, and Section~\ref{sec:experiments} outlines the experimental setup. Section~\ref{sec:results} reports the main findings, followed by qualitative and quantitative analyses in Section~\ref{sec:analysis}. Section~\ref{sec:conclusion} concludes with a discussion of limitations and future directions.

\section{Related Work}
\label{sec:related}
\paragraph{Natural Language Inference for Multilingual Evaluation.}

NLI has become a standard probe for semantic understanding in language models \cite{nighojkar-etal-2023-strong}. By requiring systems to determine whether a hypothesis follows from a premise, it offers a fine-grained test of reasoning, world knowledge, and linguistic nuance. Benchmarks such as GLUE~\cite{wang-etal-2018-glue} and SNLI~\cite{bowman-etal-2015-large} established its role in English-centric NLP, while XNLI~\cite{conneau-etal-2018-xnli} extended evaluation to 15+ languages via professional translation. Owing to its structured and interpretable format, NLI has been widely used for assessing cross-lingual transfer~\cite{heredia-etal-2024-xnlieu, bandyopadhyay-etal-2022-deep}. However, most prior work assumes monolingual evaluation—premise and hypothesis in the same language—thus overlooking mixed-lingual scenarios that are common in real multilingual discourse.

% Our work follows the tradition of NLI as a diagnostic tool but diverges in three ways: we use fully synthetic, logically controlled data; we evaluate translation consistency alongside reasoning; and we incorporate code-switching to probe multilingual alignment under conditions rarely addressed in prior studies.

\paragraph{Cross-Lingual Generalization in Large Language Models.}
Multilingual LLMs exhibit strong zero-shot transfer across languages~\cite{conneau-etal-2020-unsupervised, artetxe-etal-2020-translation}, aided by shared tokenization schemes and aligned embedding spaces. Early work with mBERT and XLM-R demonstrated cross-lingual transfer without explicit parallel training, attributed to emergent language alignment~\cite{pires-etal-2019-multilingual}. However, later studies revealed systematic biases: performance favors high-resource languages, while low-resource and morphologically rich languages often show degraded representations~\cite{schuster-etal-2019-cross-lingual}. Although recent benchmarks broaden multilingual evaluation, they typically assume monolingual inputs or perfect translation symmetry. Robustness in mixed-lingual settings—where premise and hypothesis are in different languages—remains largely untested, despite its relevance for assessing sentence-level semantic alignment beyond token overlap. Code-switching, a natural phenomenon in multilingual communities, is particularly underexplored in LLM reasoning tasks~\cite{khatri-etal-2023-translate}. Moreover, most studies use natural text, conflating syntactic variation with semantic difficulty.

Our work follows the tradition of NLI as a diagnostic tool but diverges in three ways: we use fully synthetic, logically controlled data; we evaluate translation consistency alongside reasoning; and we incorporate code-switching to probe multilingual alignment under conditions rarely addressed in prior studies.

We address this by evaluating on synthetic NLI pairs with controlled logical structure, enabling isolation of semantic consistency from linguistic noise. Our framework combines synthetic NLI data, high-quality translation, and controlled code-switching to stress-test multilingual alignment in both monolingual and mixed-lingual conditions. This design uncovers unexpected generalization patterns in instruction-tuned LLMs, challenging prevailing assumptions about cross-lingual reasoning robustness.

\begin{figure*}[ht]
\centering
\includegraphics[width=1\textwidth]{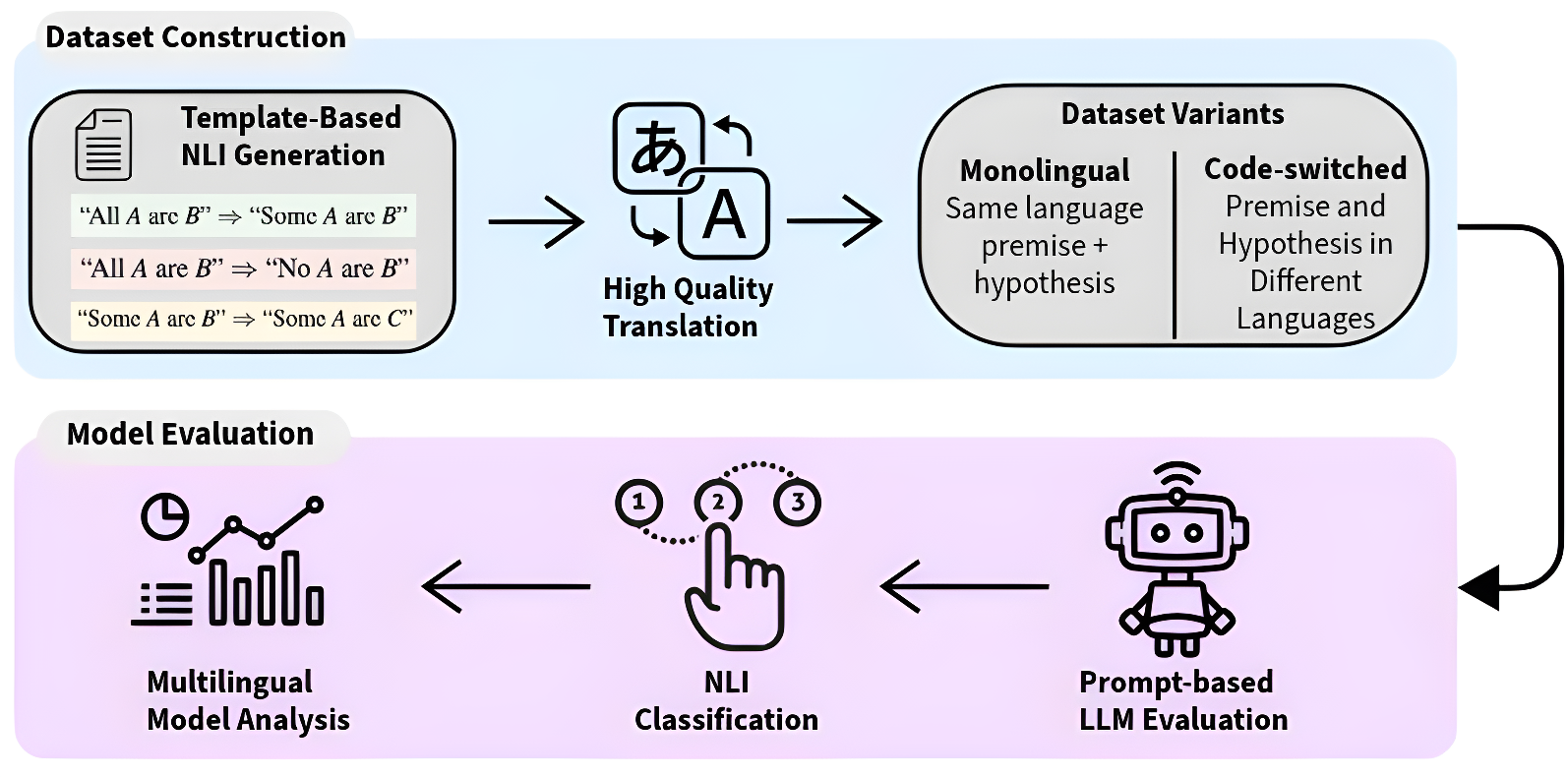}
\caption{Pipeline for Multilingual NLI Creation and Evaluation: This process involves (1) generating NLI examples using logic-based templates, (2) translating them into multiple languages with high-quality translation, (3) creating dataset variants in monolingual and code-switched formats, (4) evaluating with prompt-based LLM classification, and (5) analyzing multilingual model performance.
}
\label{fig:method}
\end{figure*}

\section{Methodology}
\label{sec:method}

This study examines the ability of LLMs to perform logically grounded NLI across languages using a controlled framework based on synthetic data generation and high-quality translation. The framework enables systematic evaluation of multilingual semantic alignment under both monolingual and mixed-lingual conditions.  Figure~\ref{fig:method} illustrates the overall methodology  for dataset construction and LLM evaluation.

\subsection{Synthetic NLI Construction}
A synthetic English NLI dataset is constructed from hand-crafted templates encoding three logical relations: entailment, contradiction, and neutrality. Each premise–hypothesis pair is derived from abstract quantifier-based patterns, with placeholders \textit{A}, \textit{B}, and \textit{C} populated using semantically coherent noun phrases to ensure plausibility. The template-based design affords precise control over compositional structure and minimizes linguistic noise, thereby isolating reasoning ability from lexical variation. Figure~\ref{fig:synthetic_templates_and_samples} presents the templates alongside example instances from the dataset.

\begin{figure*}[ht]
\centering
\footnotesize

% --- First Row: Synthetic Templates ---
\begin{minipage}{0.32\linewidth}
\begin{relcard}[Entailment]{entailbg}
``All \textit{A} are \textit{B}'' $\Rightarrow$ ``Some \textit{A} are \textit{B}''
\end{relcard}
\end{minipage}
\hfill
\begin{minipage}{0.32\linewidth}
\begin{relcard}[Contradiction]{contrabg}
``All \textit{A} are \textit{B}'' $\Rightarrow$ ``No \textit{A} are \textit{B}''
\end{relcard}
\end{minipage}
\hfill
\begin{minipage}{0.32\linewidth}
\begin{relcard}[Neutral]{neutralbg}
``Some \textit{A} are \textit{B}'' $\Rightarrow$ ``Some \textit{A} are \textit{C}''
\end{relcard}
\end{minipage}

\vspace{1em} % Space between rows

% --- Second Row: Sample Dataset Cards ---
\begin{minipage}{0.32\textwidth}
\begin{nlirelation}[Entailment]{entailbg}
\textbf{Language:} English\\[2pt]
\textbf{Premise:} All zombies are animals.\\
\textbf{Hypothesis:} Some zombies are animals.
\end{nlirelation}
\end{minipage}
\hfill
\begin{minipage}{0.32\textwidth}
\begin{nlirelation}[Contradiction]{contrabg}
\textbf{Language:} English\\[2pt]
\textbf{Premise:} All doctors are animals.\\
\textbf{Hypothesis:} No doctors are animals.
\end{nlirelation}
\end{minipage}
\hfill
\begin{minipage}{0.32\textwidth}
\begin{nlirelation}[Neutral]{neutralbg}
\textbf{Language:} English\\[2pt]
\textbf{Premise:} All monkeys are organisms.\\
\textbf{Hypothesis:} Some organisms are monkeys.
\end{nlirelation}
\end{minipage}

\caption{
\textbf{Top row:} Synthetic NLI templates encoding entailment, contradiction, and neutrality. Placeholders \textit{A}, \textit{B}, and \textit{C} are later instantiated with semantically coherent noun phrases.
\textbf{Bottom row:} Samples from the generated NLI dataset for English (en), each showing one of the three relationships: entailment (green), contradiction (red), and neutral (yellow).
}
\label{fig:synthetic_templates_and_samples}
\end{figure*}

\subsection{Multilingual Translation}
To assess inference consistency across languages, the English dataset is automatically translated into a typologically and script-diverse set of target languages using high-performance neural machine translation systems. These translations preserve the original logical relations, enabling cross-lingual evaluation under identical task structures. The selected languages—Arabic (ar), German (de), French (fr), Hindi (hi), and Swahili (sw)—cover both high- and low-resource settings and span multiple language families: Afro-Asiatic, Indo-European (Germanic, Romance, Indic branches), and Niger-Congo. Their scripts include Latin, Arabic, and Devanagari, introducing distinct orthographic and tokenization challenges. This selection also varies in morphological complexity, syntactic structure, and resource availability, providing a comprehensive basis for evaluating model robustness and cross-lingual generalization. The resulting diversity helps surface weaknesses that might remain hidden in homogeneous and high-resource-only evaluations.

\subsection{Code-Switching Probes}
To further stress-test semantic alignment, a code-switching condition is introduced in which the premise and hypothesis are presented in different languages. For each ordered pair of languages $L_1$ and $L_2$, examples are constructed with the premise in $L_1$ and the hypothesis in $L_2$, covering all possible combinations within the selected language set. This setup evaluates whether models can preserve semantic accuracy under mixed-lingual input—a common phenomenon in multilingual communication yet rarely assessed in a controlled, systematic manner.

% \subsection{Negation-Based Perturbations}

% We augment the dataset with logical perturbations designed to increase the semantic complexity. Specifically, we transform premises using negation and quantifier modification, e.g., converting ``All \textit{A} are \textit{B}'' to ``Some \textit{A} are not \textit{B}''. These transformations introduce additional inference difficulty and allow us to probe how well LLMs handle shifts in logical structure. The negated examples are also translated and code-switched, ensuring consistent evaluation across languages and configurations.
% \newpage
\subsection{Model Evaluation}

Model behavior is assessed using a prompt-based classification setup. For each example, the LLM receives a structured prompt of the form:

\begin{promptbox}[NLI Prompt Example]
\small
\textbf{Premise:} \textit{[premise]} \\[3pt]
\textbf{Hypothesis:} \textit{[hypothesis]} \\[3pt]
\textbf{Question:} Is the hypothesis entailed by the premise, contradicted by it, or unrelated? \\[3pt]
\textbf{Answer with one of:} Entailment, Contradiction, Neutral. \\[3pt]
\textbf{Answer:}
\end{promptbox}

The model outputs one of the three categorical labels. Low-temperature decoding is applied to reduce generation variability. Predictions are evaluated against gold-standard labels, and accuracy is computed across all languages and code-switching configurations.

\begin{figure*}[ht]
\centering
\includegraphics[width=1\textwidth]{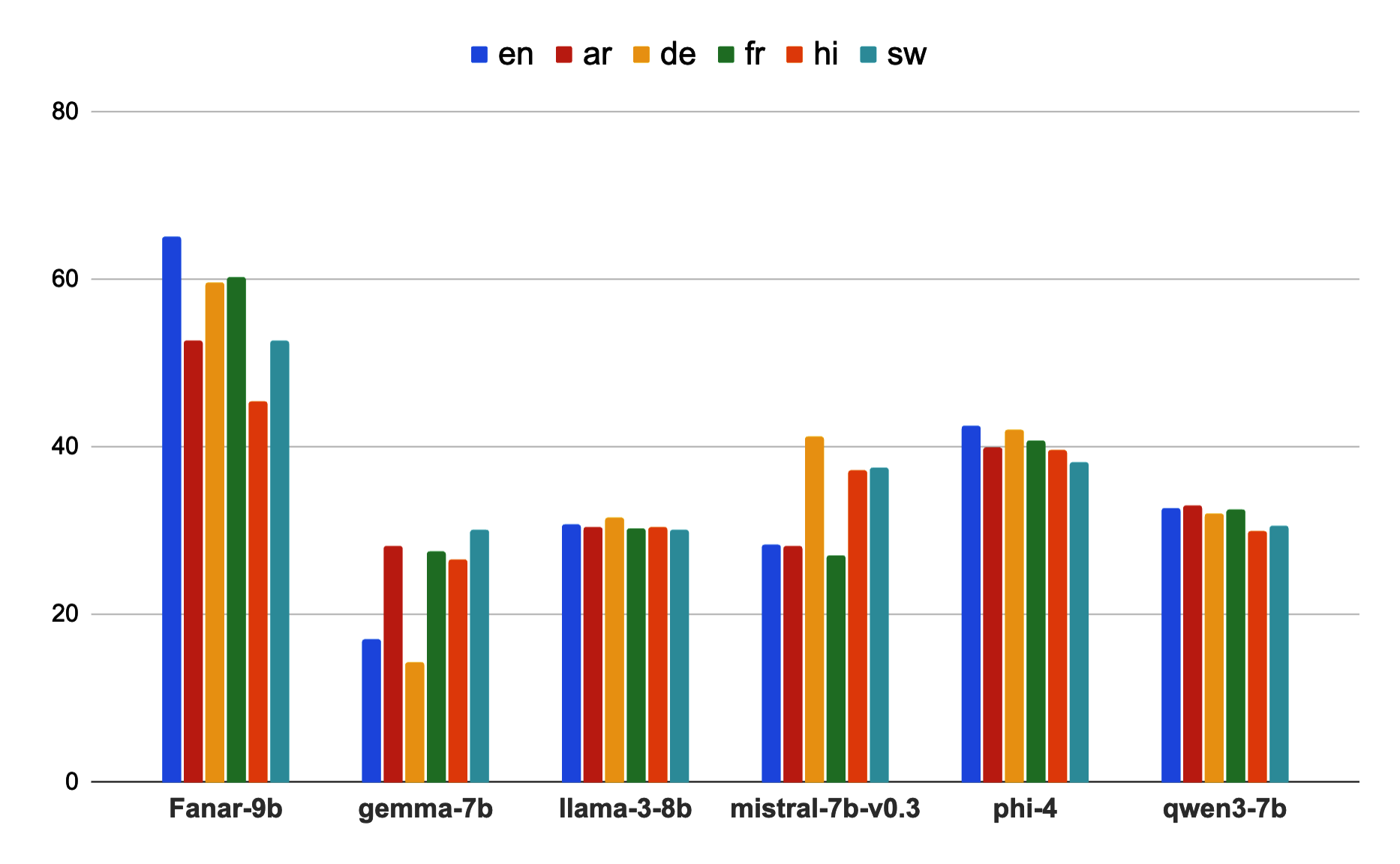}
\caption{Monolingual NLI accuracy across six languages: English (En), Arabic (Ar), German (De), French (Fr), Hindi (Hi), and Swahili (Sw); and six LLMs: Fanar-9b, Gemma-7b, Llama-3-8b, Mistral-7b-v0.3, Phi-4, and Qwen3-7b. Each bar represents the accuracy of an LLM when performing natural language inference on examples where both the premise and hypothesis are in the same language.}
\label{fig:results}
\end{figure*}
\section{Experiments}\label{sec:experiments}

\subsection{Implementation Details}

All experiments are executed using the Hugging Face Transformers library with a PyTorch backend. Inference is performed on A100 GPUs with \texttt{device\_map="auto"} enabled for memory-efficient model parallelism. Generation uses greedy decoding with a maximum of 10 new tokens per prompt to produce concise outputs while limiting hallucinations, with temperature fixed at 1.0. All models are evaluated in a zero-shot setting without task-specific fine-tuning.

\subsection{Models Evaluated}

Six multilingual instruction-tuned LLMs are evaluated, selected for diversity in architecture, size, and training data. The set includes Fanar-9B \cite{fanarteam2025fanararabiccentricmultimodalgenerative}, a multilingual model optimized for typologically diverse inputs; Gemma-7B \cite{gemmateam2024gemmaopenmodelsbased}, a decoder-only Transformer released in an instruction-tuned variant; LLaMA-3-8B \cite{grattafiori2024llama3herdmodels}, Meta’s third-generation open-weight model pretrained on a multilingual corpus; Mistral-7B-v0.3 \cite{jiang2023mistral7b}, a compact model with broad multilingual coverage; Phi-4 \cite{abdin2024phi4technicalreport}, a small but capable instruction-tuned model with strong zero-shot reasoning for its size; and Qwen3-7B \cite{yang2025qwen3technicalreport}, a multilingual model trained with extensive Chinese and non-English content. All models are evaluated using the same structured prompt format across all examples and languages to ensure comparability.

\subsection{Evaluation Scope}

The evaluation covers 36 language pairings (6×6) with 1,000 examples per pairing, balanced across the three NLI labels: \textsc{Entailment}, \textsc{Contradiction}, and \textsc{Neutral}. Both monolingual and code-switched configurations (Section~\ref{sec:method}) are included. Performance is reported as classification accuracy, computed by exact string matching between model predictions and gold standard labels.

\subsection{Reproducibility}

All experiments use publicly available model weights and reproducible scripts. The complete setup, including prompt formatting, dataset construction, translation, and inference, is implemented in Python, enabling straightforward replication and extension to additional languages and models.

\section{Results}\label{sec:results}

\subsection{Main Results}

\begin{table*}[htbp]
\centering
% Use slightly narrower columns to ensure perfect 3-across fit
\begin{tabular}{@{}m{0.305\textwidth} m{0.305\textwidth} m{0.305\textwidth}@{}}

% ============= Row 1 =============
\begin{minipage}[t]{\linewidth}
\cardtitle{Fanar-9b}
\scriptsize
\begin{tabular}{@{}lcccccc@{}}
\rowcolor{headcol}\textbf{Prem.} & En & Ar & De & Fr & Hi & Sw\\
En & \diag & \acc{51.7} & \acc{52.4} & \acc{59.4} & \acc{44.6} & \acc{43.2}\\
Ar & \acc{57.8} & \diag & \acc{61.4} & \acc{64.3} & \acc{47.2} & \acc{50.2}\\
De & \acc{53.8} & \acc{49.1} & \diag & \acc{67.0} & \acc{47.7} & \acc{46.3}\\
Fr & \acc{60.6} & \acc{51.7} & \acc{63.5} & \diag & \acc{44.7} & \acc{44.5}\\
Hi & \acc{47.4} & \acc{48.5} & \acc{43.6} & \acc{48.3} & \diag & \acc{44.6}\\
Sw & \acc{48.8} & \acc{51.6} & \acc{41.3} & \acc{54.1} & \acc{45.6} & \diag\\
\end{tabular}
\end{minipage}
&
\begin{minipage}[t]{\linewidth}
\cardtitle{Gemma-7b}
\scriptsize
\begin{tabular}{@{}lcccccc@{}}
\rowcolor{headcol}\textbf{Prem.} & En & Ar & De & Fr & Hi & Sw\\
En & \diag & \acc{29.4} & \acc{26.2} & \acc{24.8} & \acc{32.9} & \acc{26.3}\\
Ar & \acc{40.8} & \diag & \acc{39.3} & \acc{38.8} & \acc{37.4} & \acc{26.9}\\
De & \acc{30.1} & \acc{36.6} & \diag & \acc{29.7} & \acc{36.7} & \acc{24.9}\\
Fr & \acc{33.5} & \acc{40.7} & \acc{34.7} & \diag & \acc{40.1} & \acc{25.2}\\
Hi & \acc{42.6} & \acc{42.2} & \acc{43.2} & \acc{39.4} & \diag & \acc{25.4}\\
Sw & \acc{40.0} & \acc{33.6} & \acc{35.9} & \acc{41.7} & \acc{28.4} & \diag\\
\end{tabular}
\end{minipage}
&
\begin{minipage}[t]{\linewidth}
\cardtitle{Llama-3-8b}
\scriptsize
\begin{tabular}{@{}lcccccc@{}}
\rowcolor{headcol}\textbf{Prem.} & En & Ar & De & Fr & Hi & Sw\\
En & \diag & \acc{30.0} & \acc{32.0} & \acc{31.0} & \acc{29.0} & \acc{30.0}\\
Ar & \acc{30.8} & \diag & \acc{31.2} & \acc{30.5} & \acc{30.6} & \acc{29.5}\\
De & \acc{31.7} & \acc{30.2} & \diag & \acc{29.4} & \acc{30.0} & \acc{29.6}\\
Fr & \acc{30.6} & \acc{30.9} & \acc{30.5} & \diag & \acc{30.3} & \acc{31.1}\\
Hi & \acc{31.1} & \acc{31.2} & \acc{31.4} & \acc{32.7} & \diag & \acc{30.9}\\
Sw & \acc{32.1} & \acc{32.1} & \acc{31.0} & \acc{31.1} & \acc{31.5} & \diag\\
\end{tabular}
\end{minipage}
\\[0.65em]

% ============= Row 2 =============
\begin{minipage}[t]{\linewidth}
\cardtitle{Mistral-7b-v0.3}
\scriptsize
\begin{tabular}{@{}lcccccc@{}}
\rowcolor{headcol}\textbf{Prem.} & En & Ar & De & Fr & Hi & Sw\\
En & \diag & \acc{37.0} & \acc{37.7} & \acc{31.0} & \acc{32.4} & \acc{35.4}\\
Ar & \acc{36.4} & \diag & \acc{35.7} & \acc{34.8} & \acc{33.5} & \acc{34.5}\\
De & \acc{42.2} & \acc{37.6} & \diag & \acc{36.4} & \acc{35.1} & \acc{36.1}\\
Fr & \acc{34.5} & \acc{36.4} & \acc{32.9} & \diag & \acc{34.7} & \acc{34.0}\\
Hi & \acc{32.9} & \acc{34.5} & \acc{34.9} & \acc{29.5} & \diag & \acc{35.0}\\
Sw & \acc{32.5} & \acc{32.1} & \acc{32.4} & \acc{30.3} & \acc{31.9} & \diag\\
\end{tabular}
\end{minipage}
&
\begin{minipage}[t]{\linewidth}
\cardtitle{Phi-4}
\scriptsize
\begin{tabular}{@{}lcccccc@{}}
\rowcolor{headcol}\textbf{Prem.} & En & Ar & De & Fr & Hi & Sw\\
En & \diag & \acc{38.5} & \acc{38.4} & \acc{42.2} & \acc{36.6} & \acc{32.6}\\
Ar & \acc{37.3} & \diag & \acc{38.3} & \acc{39.4} & \acc{38.1} & \acc{35.1}\\
De & \acc{38.2} & \acc{38.3} & \diag & \acc{40.9} & \acc{38.5} & \acc{32.2}\\
Fr & \acc{38.0} & \acc{38.3} & \acc{38.6} & \diag & \acc{35.9} & \acc{32.8}\\
Hi & \acc{33.3} & \acc{36.8} & \acc{36.7} & \acc{37.1} & \diag & \acc{33.4}\\
Sw & \acc{35.4} & \acc{36.4} & \acc{34.3} & \acc{35.8} & \acc{33.4} & \diag\\
\end{tabular}
\end{minipage}
&
\begin{minipage}[t]{\linewidth}
\cardtitle{Qwen3-7b}
\vspace{-0.7mm}
\scriptsize
\begin{tabular}{@{}lcccccc@{}}
\rowcolor{headcol}\textbf{Prem.} & En & Ar & De & Fr & Hi & Sw\\
En & \diag & \acc{32.5} & \acc{31.7} & \acc{32.0} & \acc{29.3} & \acc{30.4}\\
Ar & \acc{32.6} & \diag & \acc{31.7} & \acc{31.4} & \acc{30.6} & \acc{31.2}\\
De & \acc{34.3} & \acc{31.8} & \diag & \acc{30.8} & \acc{30.0} & \acc{30.1}\\
Fr & \acc{32.8} & \acc{31.1} & \acc{31.7} & \diag & \acc{30.1} & \acc{31.6}\\
Hi & \acc{30.4} & \acc{31.7} & \acc{30.2} & \acc{32.3} & \diag & \acc{31.0}\\
Sw & \acc{32.8} & \acc{32.5} & \acc{32.6} & \acc{31.5} & \acc{30.9} & \diag\\
\end{tabular}
\end{minipage}
\end{tabular}
\\[0.65em]
% -------- Shared color bar (robust) --------
\vspace{0.6em}
{\scriptsize
\begin{tabular}{@{}cccccccc@{}}
\colorbox{c1}{\rule{1.4cm}{0.42cm}} &
\colorbox{c2}{\rule{1.4cm}{0.42cm}} &
\colorbox{c3}{\rule{1.4cm}{0.42cm}} &
\colorbox{c4}{\rule{1.4cm}{0.42cm}} &
\colorbox{c5}{\rule{1.4cm}{0.42cm}} &
\colorbox{c6}{\rule{1.4cm}{0.42cm}} &
\colorbox{c7}{\rule{1.4cm}{0.42cm}} &
\colorbox{c8}{\rule{1.4cm}{0.42cm}} \\
\multicolumn{8}{c}{Lower \hspace{2em} Accuracy (\%) \hspace{2em} Higher} \\
\multicolumn{1}{c}{\(<\)30} &
\multicolumn{1}{c}{30–35} &
\multicolumn{1}{c}{35–40} &
\multicolumn{1}{c}{40–45} &
\multicolumn{1}{c}{45–50} &
\multicolumn{1}{c}{50–55} &
\multicolumn{1}{c}{55–60} &
\multicolumn{1}{c}{\(\ge\)60}
\end{tabular}
}
\caption{Pairwise cross-lingual natural language inference accuracies (\%) for six language pairs
(English—En, Arabic—Ar, German—De, French—Fr, Hindi—Hi, Swahili—Sw) across six language models.
Each card presents the premise language (rows) versus the hypothesis language (columns).
Diagonal cells (\diag) indicate monolingual settings and are shaded grey, while off-diagonal cells show cross-lingual performance.
Cell colors range from light yellow (low accuracy) to dark blue (high accuracy), following the ColorBrewer YlGnBu sequential scale (legend above).}
\label{fig:codeswitch-2x3}
\end{table*}

Monolingual inference accuracy is evaluated across six languages: English (en), Arabic (ar), German (de), French (fr), Hindi (hi), and Swahili (sw). In this setting, both the premise and hypothesis are in the same language, providing a baseline measure of each model’s semantic reasoning capacity without cross-lingual interference. Results for the six evaluated LLMs are shown in Figure~\ref{fig:results}.

\paragraph{Overall Trends.}

Fanar-9B attains the highest accuracy across all languages, reaching 65.1\% in English and sustaining strong performance in lower-resource languages such as Swahili and Hindi. These results indicate a well-calibrated multilingual representation space and effective alignment of logical reasoning across typologically diverse inputs. In contrast, Gemma-7B records the lowest accuracy in nearly all languages, including 17.0\% in English and 14.3\% in German. The performance gap between Fanar-9B and Gemma-7B exceeds 40 percentage points in English, underscoring substantial differences in multilingual reasoning quality across model families.

\paragraph{Language-Specific Patterns.}

Across models, English generally achieves the highest monolingual accuracy, followed by French and German, though the magnitude of differences varies. For instance, Phi-4 performs similarly in English (43\%) and German (41\%), while LLaMA-3-8B shows minimal variance across languages, with scores clustered near 30\%. These patterns indicate that some models maintain balanced multilingual representations, whereas others exhibit pronounced bias toward high-resource and pretraining-dominant languages. Notably, Swahili, despite its lower-resource status, does not consistently underperform. In models such as Fanar-9B and Gemma-7B, Swahili accuracy is comparable to that of Indo-European languages. This outcome may reflect expanded low-resource language coverage in recent pretraining pipelines and the influence of high-quality translation data during instruction tuning.

\paragraph{Implications.}
The results reveal substantial variation in monolingual reasoning performance across languages and model architectures. While larger or more extensively instruction-tuned models often achieve higher accuracy, model size alone is not a reliable predictor; for example, LLaMA-3-8B underperforms relative to the smaller Phi-4. These patterns underscore the need to examine how training data composition, multilingual coverage, and architectural biases shape cross-lingual logical generalization, particularly for non-English and lower-resource languages.

\subsection{Code-switching}
\label{sec:results-code-switch}

The robustness of six LLMs is evaluated under code-switching conditions, in which the premise and hypothesis are presented in different languages. Table~\ref{fig:codeswitch-2x3} reports accuracy across all language pairs for each model, with off-diagonal cells representing bilingual inference. This configuration probes the ability to maintain logical consistency under mismatched linguistic inputs, a critical aspect of multilingual generalization.

\paragraph{Surprising Gains from Code-Switching.}
Several models outperform their monolingual baselines in specific code-switched configurations. For example, Gemma-7B achieves markedly higher accuracy on many bilingual pairs than on English–English (e.g., En–Hi: 32.9\% vs. En–En: 17.0\%), and Mistral-7B-v0.3 performs better on some cross-lingual inputs (e.g., Ar–En: 36.4\%) than on the corresponding monolingual cases (e.g., Ar–Ar: 28.2\%). These patterns challenge the assumption that semantic alignment necessarily degrades when models reason across linguistic boundaries.

\paragraph{Model-Specific Behaviors.}

Fanar-9B achieves the highest accuracy in both monolingual and cross-lingual settings, indicating robust multilingual alignment. In contrast, models such as Gemma-7B and Qwen3-7B display pronounced asymmetries: despite weak English monolingual performance, accuracy improves when the hypothesis is rendered in a non-English language. This pattern suggests a disproportionate reliance on hypothesis surface forms, with syntactic or lexical ambiguity in English degrading performance more than structured translations.

\paragraph{Language-Dependent Patterns.}

Accuracy gains from code-switching are unevenly distributed across languages. In several models, using Hindi, Swahili, or Arabic as the \textit{hypothesis} language yields higher performance than English, suggesting potential advantages from morphologically richer or syntactically simpler constructions in those translations. This pattern is consistent with prior findings that neural models may overfit statistical artifacts in high-resource languages, while benefiting from more literal or constrained translations in low-resource settings~\cite{coheninger2025forgetknowllmsevaluations}.

\paragraph{Implications and Hypotheses.}

The findings raise questions about the mechanisms underlying cross-lingual alignment in instruction-tuned language models. In multiple cases, accuracy is higher under code-switched conditions than in monolingual settings. Possible explanations include translation-induced lexical or syntactic variation acting as a regularization signal, improved alignment within the multilingual representation space, or simplification effects from translation. The recurrence of this pattern across diverse architectures indicates that code-switching may offer untapped potential for improving reasoning performance in multilingual applications.

\section{Cross-Lingual Analysis} \label{sec:analysis}

This section evaluates the semantic consistency of translated data and examines the representational alignment of multilingual sentences. Geometric properties of sentence embeddings are visualized across languages, and translation quality is quantified via embedding-based similarity. Given that the evaluation relies on translated versions of synthetic English inputs, verifying the preservation of semantic content across languages is essential.

\subsection{Embedding Similarity Across Translations}\label{sec:embedding-similarity}

Semantic preservation across translations is examined by visualizing sentence embeddings for five randomly selected English premise statements and their translations into six languages. Sentences are encoded with LaBSE \cite{feng-etal-2022-language} into high-dimensional vectors, then projected into three dimensions using UMAP for interpretability.

\begin{figure}[ht]
\centering
\includegraphics[width=1\columnwidth]{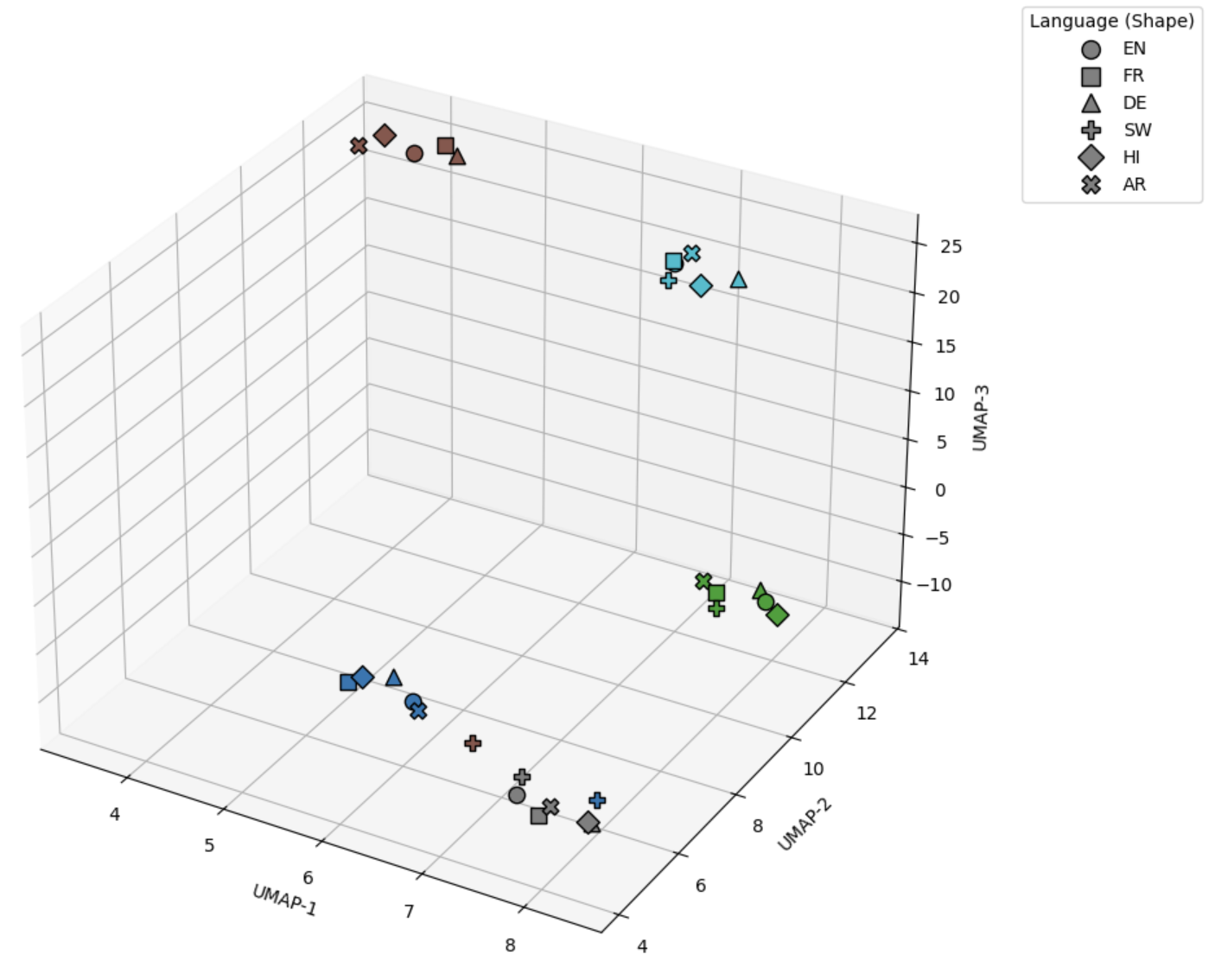}
\caption{3D UMAP projection of sentence embeddings across six languages. Each point represents a translation of one of five randomly selected NLI premise statements. Colors denote sentence identity; marker shapes indicate language (EN = English, FR = French, DE = German, AR = Arabic, HI = Hindi, SW = Swahili).}
\label{fig:embed}
\end{figure}

\paragraph{Cross-Lingual Cohesion.}
Fig.~\ref{fig:embed} shows that translations of the same sentence form tight clusters, even across typologically distant languages. This indicates high semantic consistency and suggests that the encoder maps them to similar representations despite variation in word order, morphology, or script. For instance, translations of Sentence~1 (green) remain closely grouped across all languages, supporting the preservation of intended meaning.

\paragraph{Language Variation.}
Although clusters are generally compact, certain languages display mild drift from sentence centroids. For instance, Swahili (brown in Fig.\ref{fig:embed}) shows positional deviations, likely arising from structural or morphological mismatches introduced during translation. Such patterns align with prior observations on typological variation in multilingual embedding spaces \cite{chen-etal-2025-large} and illustrate the challenge of aligning structurally divergent languages in a unified vector space. Given that the evaluation task relies on detecting fine-grained logical relations, poor or inconsistent translations could distort results. The observed cohesion across translations mitigates this concern: if translations of the same sentence consistently occupy similar embedding positions, cross-lingual performance differences are more likely to stem from genuine reasoning challenges rather than input noise.

\subsection{Translation Quality Assessment}\label{sec:translation-quality}

Semantic consistency of translations is assessed by computing cosine similarity scores between each English sentence and its translated counterpart using the LaBSE encoder, providing a direct, language-agnostic measure of semantic proximity. As shown in Table~\ref{tab:translation_quality}, similarity scores are consistently high across all languages, with French and German exhibiting the strongest alignment. Even lower-resource languages such as Swahili maintain average cosine similarities above 0.8, indicating that semantic properties are largely preserved. These results suggest that differences in inference accuracy are more likely to reflect model behavior than translation noise. Overall, the analyses confirm that the multilingual dataset preserves logical structure and meaning across languages, establishing a reliable basis for cross-lingual inference evaluation.

\begin{table}[h]
\centering
\begin{tabular}{l c r}
\toprule
\textbf{Language} & \textbf{Code} & \textbf{Avg. Cosine Similarity} \\
\midrule
French  & \texttt{fr} & \scorecell{0.912} \\
German  & \texttt{de} & \scorecell{0.895} \\
Swahili & \texttt{sw} & \scorecell{0.841} \\
Hindi   & \texttt{hi} & \scorecell{0.828} \\
Arabic  & \texttt{ar} & \scorecell{0.811} \\
\bottomrule
\end{tabular}
\caption{Semantic similarity between English premises and their translations using LaBSE embeddings (average over 100 pairs). Darker blue indicates higher similarity.}
\label{tab:translation_quality}
\end{table}

\section{Conclusion}
\label{sec:conclusion}
This study provides a controlled evaluation of multilingual semantic alignment in instruction-tuned LLMs through a synthetic, logic-based NLI framework incorporating high-quality translation and code-switching. The design isolates reasoning capabilities across languages and scripts while minimizing confounding linguistic noise. Results show that, contrary to common assumptions, reasoning performance in code-switched settings can match or exceed monolingual performance, suggesting greater robustness in cross-lingual representations than previously recognized. Translation effects may in some cases aid inference, and embedding analyses reveal strong interlingual clustering of semantically equivalent sentences, supporting the feasibility of multilingual generalization. The framework enables fine-grained probing of cross-lingual logic, identification of language-specific artifacts, and exploration of code-switching as a deliberate strategy in multilingual NLP. These findings highlight both the challenges and the opportunities for advancing reasoning-oriented multilingual evaluation.

\section{Limitations}
\label{sec:limitation}
\paragraph{Synthetic Nature of the Dataset.}
The use of synthetic NLI examples enables precise control over logical form and compositional structure but may limit ecological validity. The templates, while semantically well-formed, cannot fully capture the diversity and ambiguity of natural multilingual discourse. Consequently, performance on these tasks may not directly translate to real-world reasoning ability. Future work could mitigate this limitation by supplementing template-based data with linguistically diverse or naturally occurring sentences, curated and verified across languages to preserve logical consistency.

\paragraph{Reliance on Machine Translation.}
The evaluation of cross-lingual alignment assumes that machine translation preserves the intended semantics of the original English examples. Neural translation systems—particularly for low-resource languages—can introduce meaning shifts, simplifications, or structural divergences that alter the logical relationship between premise and hypothesis. Although state-of-the-art translation models were used and their quality assessed (Section~\ref{sec:analysis}), residual errors may still influence downstream reasoning. Future extensions could incorporate human verification of a subset of translations or employ multilingual LLMs to produce language-native examples directly, avoiding translation as an intermediate step.

% \newpage
\nocite{*}
\section{Bibliographical References}\label{sec:reference}

\bibliographystyle{lrec2026-natbib}

% \section{Language Resource References}
% \label{lr:ref}
% \bibliographystylelanguageresource{lrec2026-natbib}
% \bibliographylanguageresource{languageresource}

\end{document}